\documentclass[lettersize,journal]{IEEEtran}
\usepackage{amsmath,amsfonts}
\usepackage{algorithm}
\usepackage{algpseudocode}
\usepackage{array}
\usepackage[caption=false,font=normalsize,labelfont=sf,textfont=sf]{subfig}
\usepackage{textcomp}
\usepackage{stfloats}
\usepackage{url}
\usepackage{verbatim}
\usepackage{graphicx}

\usepackage {adjustbox}
\usepackage{booktabs}
\usepackage{enumitem}
\usepackage{multirow}
\usepackage{colortbl}
\usepackage{xcolor}
\usepackage[normalem]{ulem}
\usepackage{amssymb} 

\newtheorem{theorem}{Theorem}

\newtheorem{theorem*}{Theorem}

\newcommand{\Input}{\Require}

\def\BibTeX{{\rm B\kern-.05em{\sc i\kern-.025em b}\kern-.08em
    T\kern-.1667em\lower.7ex\hbox{E}\kern-.125emX}}
\usepackage{balance}

\usepackage{xr-hyper}
\begin{document}
\title{Hidden Human-Like Nature of Machine-Generated Texts: Theory and Detection Enhancement}

\author{Chenwang~Wu,
        Yiu-ming~Cheung,~\IEEEmembership{Fellow,~IEEE,}
        Bo~Han,~\IEEEmembership{Senior Member,~IEEE,}
        and Defu Lian
\IEEEcompsocitemizethanks{\IEEEcompsocthanksitem Chenwang Wu, Yiu-ming Cheung, and Bo Han are affiliated with the Department of Computer Science, Hong Kong Baptist University, Hong Kong, China.\protect\\
E-mail: \{cscwwu, ymc, bhanml\}@comp.hkbu.edu.hk.
\IEEEcompsocthanksitem Defu Lian is affiliated with the School of Computer Science and Technology, University of Science and Technology of China, Hefei, Anhui
230000, China. He is also affiliated with the State Key Laboratory of Cognitive Intelligence.\protect\\
E-mail: liandefu@ustc.edu.cn.
\IEEEcompsocthanksitem Corresponding author: Yiu-ming~Cheung.}% <-this % stops an unwanted space
}

\markboth{Hidden Human-Like Nature of Machine-Generated Texts: Theory and Detection Enhancement}%
{How to Use the IEEEtran \LaTeX \ Templates}

\maketitle

\begin{abstract}

Machine-generated texts (MGTs) produced by large language models (LLMs) are increasingly prevalent across various applications, while their potential misuse in fake news propagation and phishing has raised serious concerns, highlighting the need for MGT detection. Existing paragraph-level detection methods commonly treat MGTs as entirely machine-like, overlooking the hidden human-like nature of machine-generated texts: even fully machine-generated texts may contain spans that are highly consistent with human writing. To this end, we first reveal the existence of such hidden human-like spans, and then theoretically analyze their impact on detection. Our analysis shows that these spans increase the sentence complexity for detection, thereby making MGT detection intrinsically harder. Based on this finding, we propose a model-agnostic stacked enhancement framework that improves existing detectors by reducing the influence of hidden human-like spans. Specifically, we model span-level retention decisions as a latent-variable problem and instantiate the optimization with a hard-EM-inspired procedure, where the detector iteratively filters confidently human-like subsequences and refines itself on the remaining text. Extensive experiments across various LLMs and practical scenarios demonstrate that the proposed framework consistently enhances existing detectors. Notably, the framework can also work in a training-free manner, offering flexibility and scalability for practical deployment.

\end{abstract}

\begin{IEEEkeywords}
Machine-generated text detection, model-based detection, hidden human-like spans
\end{IEEEkeywords}

\section{Introduction}

\IEEEPARstart{T}{he} rapid development of large language models (LLMs) \cite{achiam2023gpt,radford2019language} has led to a surge in machine-generated texts (MGTs). Although MGTs can support a wide range of applications, their potential misuse has raised serious concerns about fake news propagation \cite{zellers2019defending}, phishing \cite{hong2012state}, and academic fraud \cite{alshurafat2024factors}. For example, cybercriminals can create realistic phishing emails to commit fraud. These risks highlight the need for effective MGT detection to ensure safe and transparent AI systems \cite{ciftci2020fakecatcher}. 
Sentence-level detection is inherently challenging, as individual sentences often contain insufficient contextual cues and can be easily confused with human-written text. Therefore, this paper focuses on paragraph-based detection, which can better leverage contextual information to achieve robust detection \cite{tulchinskii2024intrinsic}.

Existing detection methods can be roughly separated into two lines:
(1) Feature-based detection methods identify MGT by using distinctive properties of generated text, e.g., output log probability \cite{mitchell2023detectgpt,solaiman2019release}, objectivity and sentiment of the language \cite{guo2023close}, cross entropy \cite{guo2024biscope}, and intrinsic dimensions \cite{tulchinskii2024intrinsic}. However, due to the complexity of textual data, manually-extracted features from limited data often fail to fully capture intricate patterns and structures, thus leading to poor generalization across generative models.
(2) Model-based detection methods use entire texts as inputs, allowing detectors to implicitly learn distinguishing features during training. Such methods are more flexible than feature engineering methods and have received increasing attention. Classic examples include energy-based models \cite{tulchinskii2024intrinsic}, small language models \cite{mireshghallah2023smaller}, LLMs \cite{verma2024ghostbuster}, and graph neural networks \cite{zhong2020neural}. Besides, the quality of data representation is crucial for learning detection models, such as those that use pre-trained text features \cite{crothers2022adversarial} and probability lists from open-source LLMs \cite{wang2023seqxgpt}.

Usually, the methods mentioned above commonly treat machine-generated texts as entirely machine-like. This simplified assumption, however, overlooks the hidden human-like nature of MGTs: even a fully machine-generated text may contain spans that are highly consistent with human writing. These human-like spans weaken the discriminative evidence available to detectors and make the boundary between human-written and machine-generated texts less separable. Under the circumstances, three key research questions naturally arise:
\begin{itemize}[leftmargin = *]
    \item \textbf{RQ1}: Do fully machine-generated texts commonly contain hidden human-like spans that are consistent with human writing?
    \item \textbf{RQ2}: RQ2: If the answer to RQ1 is yes, how do such hidden human-like spans affect MGT detection, and what benefits can be obtained by handling them?
    \item \textbf{RQ3}: For the challenges of RQ2, how can we refine the detection model to overcome them?
\end{itemize}
This paper aims to study MGT detection by solving these three questions.

\begin{figure}[t]
	\centering
	\includegraphics[width=1.\linewidth]{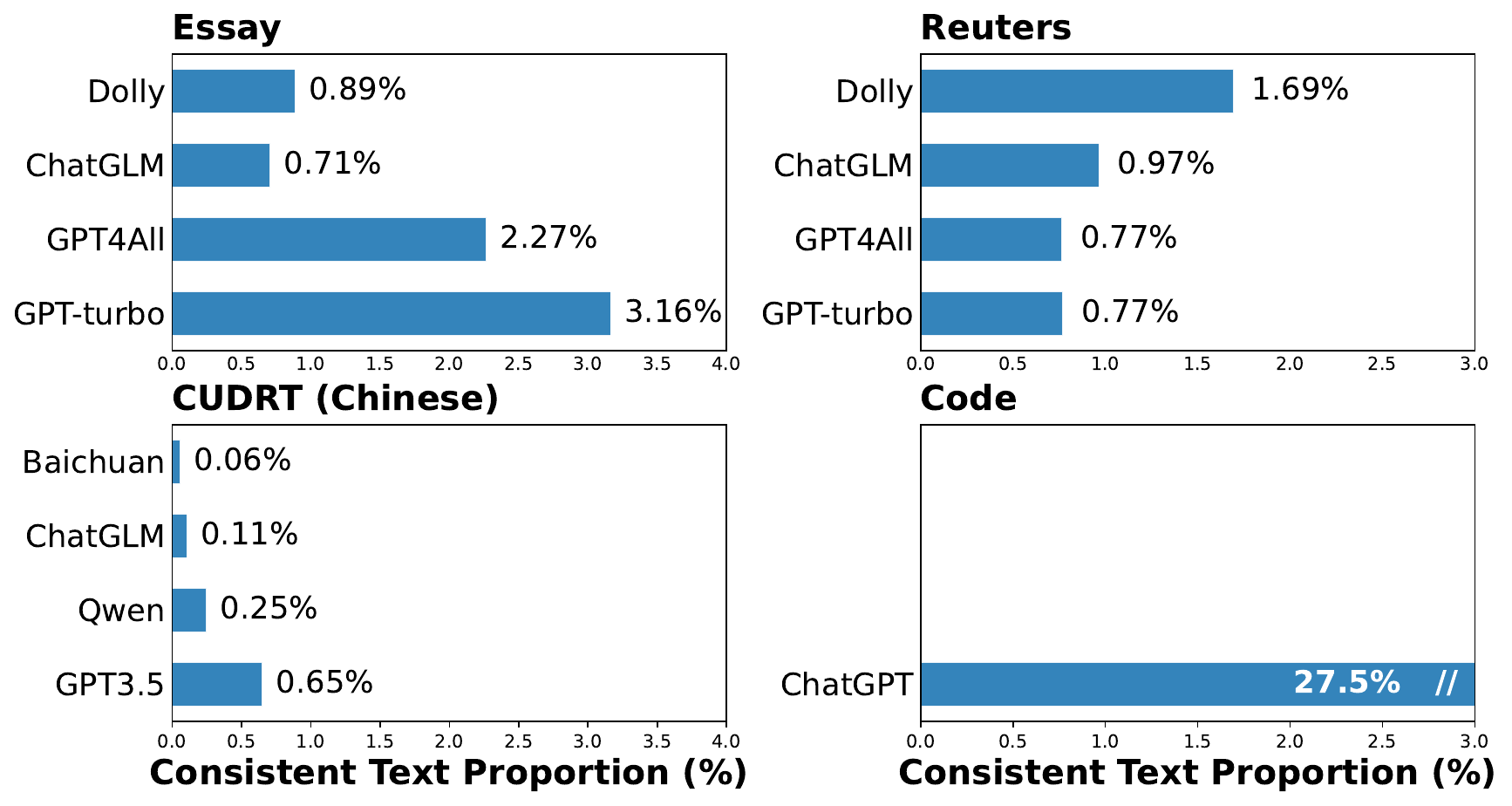}
 \vspace{-0.5cm}
	\caption{The proportion of consistent sentences between humans and LLMs. Although some proportions are small, non-zero values across datasets provide empirical evidence for the presence of hidden human-like spans.}
	\label{fig: motivation}
    \vspace{-0.4cm}
\end{figure}
First, unlike the work \cite{zhang2024llm, wang2024llm} that focuses on explicit mixed text produced through human-machine collaboration, we investigate a more implicit phenomenon: even seemingly pure MGTs can contain hidden human-like spans (\textbf{RQ1}, Section \ref{sec: exist_mix}). Specifically, even if the text is fully generated by an LLM rather than collaboratively written by humans and machines, some of its spans can still be highly consistent with human writing. Fig. \ref{fig: motivation} shows the proportion of consistent sentences in machine and human text across four datasets. This proportion provides a conservative lower bound for the existence of hidden human-like spans; a value greater than 0 indicates that such spans are present (more discussion is in Section \ref{sec: exist_mix}). These results suggest that fully machine-generated texts are not necessarily entirely machine-like. Since exact matching provides only a conservative lower bound, the actual occurrence of hidden human-like spans may be underestimated.

The presence of hidden human-like components in MGTs requires us to analyze their impact on detection (\textbf{RQ2}, Section \ref{sec: motivation_theory}). Therefore, we theoretically reveal the sentence complexity required by the best possible detector to achieve a given AUROC. The results indicate that the required complexity increases with the proportion of human-like spans within MGTs, and reaches its minimum when this proportion is 0. This implies that the hidden human-like nature of MGTs inherently hinders detection.

To tackle the issue (\textbf{RQ3}, Section \ref{sec: methodology}), we first derive a theory-motivated filtering principle and then operationalize it as a practical stacked enhancement framework. Specifically, since the span-level retention decisions are unobserved in practice, we introduce latent retention variables to determine whether each sequence should be retained for detection or filtered as confidently human-like. This formulation naturally leads to a hard-EM-inspired optimization strategy, where latent mask estimation and detector refinement are alternately performed within a practical stacked enhancement framework. In the hard E-step, the detector filters out confidently human-like sequences, yielding a simpler post-filter distribution. In the hard M-step, the same detector is refined using the remaining sequences. The improved detector is then reused in the next E-step, enabling iterative self-enhancement.

Our contributions are summarized as follows.
\begin{itemize}[leftmargin=*]
\item \textbf{We reveal the hidden human-like nature of machine-generated texts (Section \ref{sec: exist_mix}).} 
We show that even fully machine-generated texts may contain human-like spans that are difficult to distinguish from human writing. This reveals a more general and overlooked property of MGTs, making detection more challenging than previously recognized.
\item \textbf{We theoretically analyze the impact of hidden human-like components on detection (Section \ref{sec: motivation_theory}).} 
We provide theoretical evidence that human-like components within MGTs can increase the difficulty of detection. This analysis explains prior empirical observations and motivates explicitly modeling such hidden components.
\item \textbf{We propose a theory-motivated detection enhancement framework (Section \ref{sec: methodology}).} 
Motivated by the theoretical finding, we first establish a filtering-based enhancement principle and then instantiate it as a practical stacked framework by modeling span-level retention decisions as latent variables. The key contribution is a general framework that can also support other inference and optimization strategies.
\item \textbf{Extensive experiments demonstrate the effectiveness of the proposed detection enhancement framework (Section \ref{sec: experiment}).} 
We conduct experiments across real-world scenarios, including cross-LLM, cross-domain, paraphrased texts, and explicit mixed scenarios, and the results demonstrate our framework's potential to enhance existing detectors.
\end{itemize}

\section{Related Work}
\label{sec: related}

In this section, we provide an overview of the detection efforts, which can be categorized into watermark-based, feature-based, and model-based methods \cite{fraser2025detecting}.

The watermarks hidden in the text indicate it is AI-generated. Existing work \cite{kirchenbauer2023watermark} suggested mildly promoting the use of predefined "green" tokens during text generation and proposed a statistical test method to detect watermarks. Unigram-Watermark \cite{zhao2023provable} was proposed by extending existing approaches with a simplified fixed grouping strategy. However, watermark-based methods require high privileges over the model, which limits their wide applicability. Furthermore, recent work \cite{zhang2024watermarks, jovanovic2024watermark} has shown that strong watermarking is difficult to achieve and is vulnerable to watermark attacks.

Feature-based methods exploit statistical and linguistic characteristics to distinguish human-written and machine-generated texts without training on specific datasets. Early studies primarily use token-level probability statistics, such as Log-Likelihood \cite{solaiman2019release}, Entropy \cite{gehrmann2019gltr}, and their variants \cite{su2023detectllm,zhouadadetectgpt}, based on the observation that generated text often exhibits distinctive probability patterns. DetectGPT \cite{mitchell2023detectgpt} further observes that small perturbations of generated texts usually lead to lower log probabilities than the original samples. Subsequent works extend this idea with more efficient or task-specific perturbation, rewriting, continuation, and comparison strategies \cite{bao2024fast,yangdna2024,yang2023zero,nguyen2024simllm,zhou2026learn}.
Beyond probability-based scores, recent studies explore richer intrinsic signals of generated texts, including intrinsic dimensionality \cite{tulchinskii2024intrinsic}, surrogate-model activation features \cite{chen2025repreguard}, temporal probability patterns \cite{xutraining}, relative probability spectra \cite{xu2024detecting}, and uncertainty in style perception \cite{wu2025moses}. More recent methods further consider hierarchical relations, repair effort, out-of-distribution generalization, prompt inference, and multi-level linguistic distributions \cite{hedetree,zhudna,zenghuman,chen2025ipad,guohld}. Although these methods reveal diverse properties of MGTs, they often rely on specific observable features or model-dependent signals, making it difficult to comprehensively capture the complex, hidden human-like characteristics of generated texts.

Model-based methods represent a classical paradigm for MGT detection, in which detectors are trained as binary classifiers on human-written and machine-generated text. Unlike explicit feature-based methods, model-based detectors usually take the entire text as input and learn implicit discriminative patterns \cite{bengio2013representation} during training. Representative systems, such as OpenAI Detector \cite{solaiman2019release}, ChatGPT Detector \cite{guo2023close}, GPTZero \cite{gptzero}, and G3 Detector \cite{zhan2023g3detector}, collect texts generated by different LLMs to train unified classifiers. Recent studies further show that smaller or partially trained language models can serve as effective general detectors \cite{mireshghallah2023smaller}, and that detectors should be evaluated against increasingly advanced generation models \cite{pagnoni2022threat}.
Beyond raw-text classification, some methods design alternative inputs or auxiliary signals. GLTR \cite{gehrmann2019gltr} uses token-rank features for logistic regression, while SeqXGPT \cite{wang2023seqxgpt} treats logit sequences as waveform-like signals. LLMDet \cite{wu2023llmdet} incorporates surrogate-model perplexity, and graph-based methods model factual structures to capture document-level consistency \cite{zhong2020neural}. More recent studies explore advanced training paradigms, including mixed decoding strategies in GPT-Pat \cite{yu2023gpt}, positive-unlabeled learning in MPU \cite{tianmultiscale}, adversarial training in RADAR \cite{hu2023radar}, surrogate-model training for unknown-source detection in Ghostbuster \cite{verma2024ghostbuster}, and multimodal contrastive learning in DGM$^4$ \cite{shao2024detecting}.
Although these methods achieve strong supervised performance, their generalization often decreases when the testing distribution differs from the training data \cite{liang2023gpt,ippolito2020automatic, wu2023influence}. For example, detectors may misclassify texts written by non-native English speakers \cite{liang2023gpt}; some categories, such as recipes, are easier to detect than stories or news \cite{tulchinskii2024intrinsic}. These limitations suggest that supervised model-based detectors may overfit to specific generators, domains, or decoding patterns, and thus remain vulnerable to hidden human-like characteristics in MGTs.

\section{Hidden Human-Like Spans Make Detection Harder}

In this section, we first demonstrate that machine-generated texts may contain hidden human-like spans. We then formalize the conventional MGT detection setting and theoretically analyze how such spans increase the difficulty of detection.

\subsection{Existence of Hidden Human-Like Spans}

\label{sec: exist_mix}

Most detection strategies implicitly assume that text is generated homogeneously by either humans or machines, and therefore take the entire text as input for detection. Recent studies \cite{zeng2024detecting, zhang2024llm} have begun to consider explicit human-machine mixed texts, where a document is collaboratively written or edited by humans and LLMs. In contrast, this paper focuses on a subtler, more easily overlooked phenomenon: even when a text is fully generated by an LLM, some of its spans may still be highly consistent with human writing.

We refer to this phenomenon as the \textbf{hidden human-like nature of machine-generated texts}. Specifically, modern LLMs, owing to their strong generative capability and exposure to large-scale human-written corpora, may produce spans that closely match human-written content. Typical examples include simple sentence structures (e.g., ``Hello World''), fixed-format expressions (e.g., ``Thank you for your letter''), and common factual or template-like patterns (e.g., specific places, names, or conventional descriptions). Although these spans are produced by machines, they exhibit human-like characteristics and thus provide weak or even misleading evidence for machine authorship. This suggests that fully machine-generated texts are not necessarily entirely machine-like.

To empirically verify the existence of such hidden human-like spans, we calculate the proportion of consistent sentences between LLM-generated and human-written texts. Although this proportion is only a conservative proxy for the true proportion of human-like spans, it establishes a lower bound: a non-zero value indicates that at least some hidden human-like spans exist (detailed discussion can be found in Section II of the Supplementary Material \footnote{\url{https://github.com/Daftstone/MGTD/blob/master/Supplementary_Material.pdf}}). More complex statistical tests may better approximate the true proportion, but they are unnecessary for our purpose, since our goal here is not to precisely estimate this proportion, but to demonstrate the existence of hidden human-like spans. As shown in Fig. \ref{fig: motivation}, although the observed proportions are generally small due to the limited scale of the available human and machine texts, their nonzero values consistently indicate the presence of hidden human-like spans across different datasets. For theoretical convenience, we refer to the machine-generated texts containing such hidden human-like spans as mixed texts in the following analysis, unless otherwise specified.

\subsection{Formalization of MGT Detection}
\label{sec: preliminary}

Before analyzing the detection challenge introduced by hidden human-like spans, we first formalize the conventional MGT detection setting.

\textbf{Text Data Definition}. Following the existing definition \cite{chakrabortyposition}, let $\mathcal{S}$ denote the sentence space. We define the human sentence distribution as $h(s)$ for $s\in\mathcal{S}$, and similarly define the machine one as $m(s)$. This allows us to model texts containing multiple sentences under both IID and non-IID settings.

\begin{itemize}[leftmargin = *]
\item\textbf{Sentence IID Setting}. If a text $S$ contains $n$ sentences, denoted as $S:=\{s_i\}_{i=1}^n$, and each sentence $s_i$ is i.i.d. drawn from either the human distribution $h(s)$ or the machine distribution $m(s)$, then the human-written text can be denoted as $S\sim h^{\otimes n}(s)$, while the machine-generated text can be denoted as $S\sim m^{\otimes n}(s)$, where $h^{\otimes n}:=h\otimes h\otimes \ldots \otimes h$ ($n$ times), and $m^{\otimes n}$ is defined similarly.
\item\textbf{Sentence Non-IID Setting}. We follow a practical setting \cite{chakrabortyposition, loureiro2024topics} for various language tasks. Assume that $\rho$ characterizes the dependency strength between sentences, and $\sum_i s_i$ represents the topic consistency of these sentences, where each $s_i$ is embedded in a semantic vector space. For instance, if the preceding text discusses hospitals and doctors, then the topic consistency of these sentences would point to the medical field.
If $T_i$ denotes the random variable corresponding to the $i$-th sentence, then the semantics of the current sentence are determined by both historical topic consistency and the language distribution itself:
\end{itemize}
\[
\mathbb{E}\left[T_i \mid T_{i-1}=s_{i-1}, \cdots, T_1=s_1\right]
=\rho \frac{\sum_{k=1}^{i-1} s_k}{i-1}+(1-\rho) \mathbb{E}\left[T_i\right].
\]
When $\rho=0$, this non-IID setting degenerates into the IID setting. As $\rho$ increases, the current sentence becomes more dependent on previous sentences.

\textbf{Problem Definition}. The task of MGT detection can be formulated as a binary classiﬁcation problem. The detector $f$ maps the text $S$ to a real value $f(S)\in[0,1]$, which indicates the confidence of machine generation. If $f(S)$ is greater than a predefined threshold $r$, text $S$ is predicted to be machine-generated; otherwise, it is human-generated. Assuming that text $S$ contains $n$ sentences, existing work \cite{chakrabortyposition} has proven that the likelihood-ratio-based detector can achieve the upper bound of AUROC and is the best possible detector:
\begin{equation}
    f^*(S):= \begin{cases}\text { Machine Text} & \text { if \ \ } m^{\otimes n}(S) \geq h^{\otimes n}(S), \\ \text { Human Text} & \text { if \ \ } m^{\otimes n}(S)<h^{\otimes n}(S) .\end{cases}
    \nonumber
\end{equation}
This formulation corresponds to the conventional homogeneous view of MGT detection, in which a text is modeled as drawn entirely from either the human or the machine distribution. In the following sections, we relax this view by considering the hidden human-like nature of MGTs, in which fully machine-generated texts may still contain spans that are highly consistent with human writing.

\subsection{Detection Challenge of Hidden Human-Like Spans}
\label{sec: motivation_theory}

The existence of hidden human-like spans requires us to revisit the conventional formulation of MGT detection. Although generated by machines, these spans are highly consistent with human writing and can thus be modeled as human-distribution components from the detection perspective. This subsection theoretically analyzes how they affect the difficulty of detection.

Building on the formalization in Section \ref{sec: preliminary}, we consider a machine-generated text $S:=\{s_i\}_{i=1}^n$ with $n$ sentences. Let $\alpha$ denote the proportion of hidden human-like spans within the MGT. For theoretical analysis, these spans are modeled as being drawn from the human sentence distribution $h(s)$, while the remaining machine-indicative spans are drawn from the machine sentence distribution $m(s)$. Without loss of generality, assume that $(1-\alpha)n$ sentences $\left\{s_i\right\}_{i=1}^{(1-\alpha)n}$ are from $m(s)$ and $\alpha n$ sentences $\left\{s_i\right\}_{i=(1-\alpha)n+1}^{n}$ are from $h(s)$. The human-written text $S$ consists of sentences i.i.d. drawn from $h(s)$. Therefore, the distribution of MGTs with hidden human-like spans is denoted as $S\sim m^{\otimes (1-\alpha)n}h^{\otimes \alpha n}(s)$, denoted as $M(S)$, while the human-written text follows $S\sim h^{\otimes n}(s)$, denoted as $H(S)$. Then, the best possible detector under this setting is:
\begin{equation}
    f^*(S):= \begin{cases}\text{Machine Text}& \text { if } M(S) \geq H(S), \\ \text{Human Text}& \text { if } M(S) < H(S).\end{cases}
    \nonumber
\end{equation}
For the non-IID setting, assume that text $S$ contains $L$ independent sequences $\{v_i\}_{i=1}^L$, where each sequence $v_i$ consists of $c_i$ dependent sentences. This assumption is reasonable due to factors such as topic independence and context switching. We then derive the sentence complexity bound of MGT detection as follows.

\begin{theorem}[\textbf{Sentence Complexity of Mixed Text Detection under Non-IID Setting}]
\label{theorem: complexity_noniid}
Assume the total variation distance between the human and machine distributions is $TV(m,h)=\delta>0$. Let the text contain $n$ sentences, with $\alpha$ representing the proportion of hidden human-like spans within the MGT. To achieve an AUROC of $\epsilon$, the required sentences $n$ for the best possible detector is:
\begin{equation}
    \begin{aligned}
& n=\Omega\left(\frac{1}{(1-\alpha)^2\delta^2}\ln\left(\frac{1}{1-\epsilon}\right)+\frac{1}{(1-\alpha)\delta}\sum_{j=1}^L(c_j-1)\rho_j\right. \\
& \left. +\left(\frac{1}{(1-\alpha)^3\delta^3} \left(\sum_{j=1}^L\left(c_j-1\right) \rho_j\right)\ln \left(\frac{1}{1-\epsilon}\right)\right)^{1/2}\right).
\nonumber
\end{aligned}
\end{equation}
\end{theorem}

The proof is given in Section IV of the Supplementary Material, and the IID result is provided in Section III-A. This theorem shows that achieving better detection performance, i.e., a larger $\epsilon$, requires higher sentence complexity $n$, aligning with existing findings \cite{kirchenbauerreliability}. More importantly, the required sentence complexity increases with the proportion $\alpha$ of hidden human-like spans. When $\alpha=0$, i.e., the MGT contains no hidden human-like spans and is entirely machine-like from the distributional perspective, the detector has the lowest complexity bound. Therefore, hidden human-like spans inherently hinder MGT detection, and reliable detection becomes easier when fewer such spans are present. Empirical evidence for this result is given in Section \ref{sec: mixed_degree}.

\section{Proposed Method}
\label{sec: methodology}

The theoretical results in Section \ref{sec: motivation_theory} provide a direct design principle: since hidden human-like spans increase the sentence complexity of detection, reducing their influence should make detection easier. Ideally, removing a portion of such spans from MGTs would decrease the effective proportion $\alpha$ and improve detection. However, this idea faces two practical challenges: (1) the text-level label is unknown during detection, and (2) the span-level retention status, i.e., whether each span should be filtered or retained for detection, is unobserved.

To address these challenges, Section \ref{sec: theoretical_improvemenets} theoretically establishes the conceptual principle: if a small portion of human-like spans can be filtered from all texts in a label-agnostic manner, the effective sentence complexity of MGT detection can be reduced.  However, Section \ref{sec: theoretical_improvemenets} still relies on an idealized filtering operation that can select a small proportion of human-like spans, which is not directly available in practice.  Section \ref{sec: stacked_framework} therefore provides a practical instantiation of this principle by modeling span-level retention decisions as latent variables and estimating the corresponding retention masks through a hard-EM-inspired stacked procedure.

\subsection{Theory-motivated Label-agnostic Filtering Principle}
\label{sec: theoretical_improvemenets}

We first address the first challenge, i.e., the text-level label is unknown during detection. Instead of assuming access to MGT labels, we consider a compromise label-agnostic filtering strategy that removes a small proportion of human-like spans from all texts. Under the setting of Section \ref{sec: motivation_theory}, suppose that we filter an $\alpha_s$ proportion of human-like spans from all texts, where $\alpha_s<\alpha$. These filtered spans are not necessarily human-written; rather, they are distributionally close to human writing and are thus modeled by $h(s)$ from the detection perspective. The following result shows that such filtering can reduce the sentence complexity required for reliable detection.

\begin{theorem}[\textbf{Sentence Complexity of Label-Agnostic Filtering-based Method under Non-IID Setting}]
\label{theorem: our_complexity_noniid}
Consider the MGT detection under the assumption of Theorem \ref{theorem: complexity_noniid}. If we filter an $\alpha_s$ ($<\alpha$) proportion of hidden human-like spans from all texts, then to achieve an AUROC of $\epsilon$, the required sentences $n$ for the best possible detector is
\begin{equation}
    \begin{aligned}
& n=\Omega\left(\frac{1-\alpha_s}{(1-\alpha)^2\delta^2}\ln\left(\frac{1}{1-\epsilon}\right)+\frac{1}{(1-\alpha)\delta}\sum_{j=1}^L(c_j-1)\rho_j\right. \\
& \left. +\left(\frac{1-\alpha_s}{(1-\alpha)^3\delta^3} \left(\sum_{j=1}^L\left(c_j-1\right) \rho_j\right)\ln \left(\frac{1}{1-\epsilon}\right)\right)^{1/2}\right).
\end{aligned}
\nonumber
\end{equation}
\end{theorem}

Compared with Theorem \ref{theorem: complexity_noniid}, Theorem \ref{theorem: our_complexity_noniid} reduces the effective text-complexity terms by the factor $1-\alpha_s$. This indicates that filtering hidden human-like spans can reduce the sentence complexity required to achieve the same AUROC, that is, enhancing detection. When $\alpha_s=0$, i.e., no filtering is performed, Theorem \ref{theorem:  our_complexity_noniid} degenerates into Theorem \ref{theorem: complexity_noniid}. The IID result is provided in Section III-B of the Supplementary Material, where similar findings are obtained.

Although this strategy resolves the first challenge through label-agnostic filtering, it still assumes that the spans to be filtered can be determined. In practice, the optimal span retention mask is unobserved. The next subsection, therefore, instantiates this conceptual strategy as a practical framework by modeling span-level retention decisions as latent variables.

\begin{figure*}[t]
	\centering
	\includegraphics[width=0.95\linewidth]{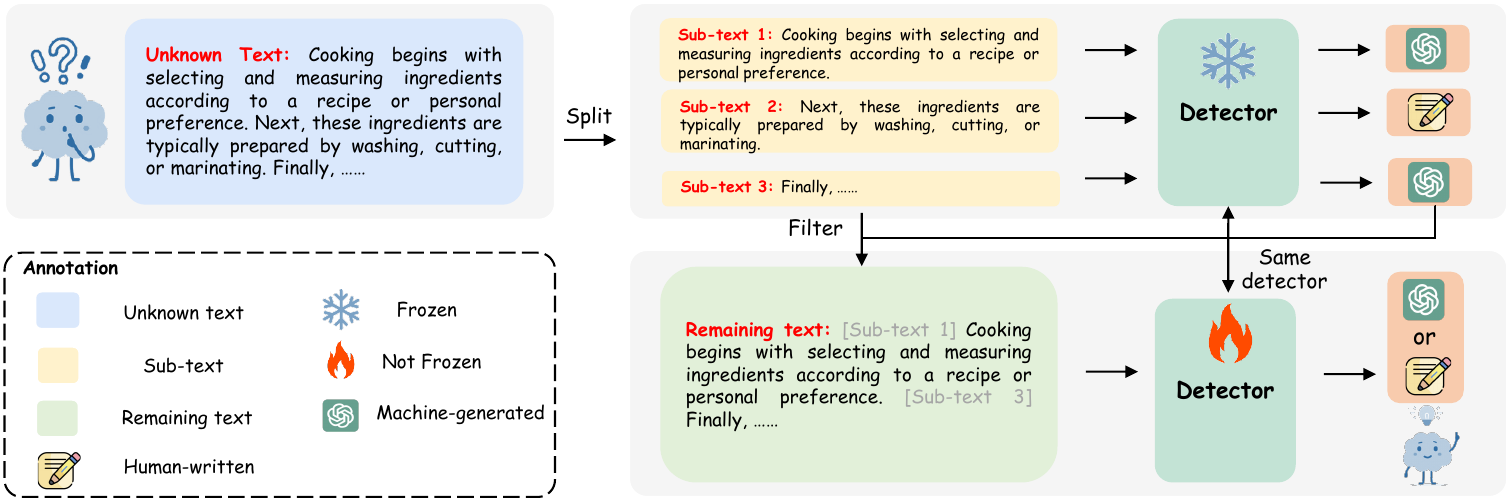}
 % \vspace{-0.05cm}
	\caption{The inference process of the proposed framework. In the filtering step (top-right), the unknown text is split into sub-sequences. The trained detector runs on each sub-sequence to estimate a latent retention mask, where confidently human-like sub-sequences are filtered out; (2) In the detection step (bottom-right), the remaining text is concatenated and fed back into the exact same detector. The output of this second pass is the final prediction.}
	\label{fig: framework}
    \vspace{-0.2cm}
\end{figure*}

\subsection{Instantiation via Latent-variable Stacked Optimization}
\label{sec: stacked_framework}

Within the conceptual strategy, we now address the second challenge: the optimal retention decision for each span is unobserved in practice. To this end, we formulate span-level retention as latent variables and derive a tractable, hard-EM-inspired optimization procedure. In this procedure, the hard E-step estimates a latent retention mask that determines which spans should be retained for detection, and the hard M-step optimizes the detector based on the retained text. This is instantiated as the stacked framework shown in Fig. \ref{fig: framework}: the first pass estimates the latent retention mask by conservatively filtering confidently human-like spans, while the second pass applies the same detector to the retained text for final detection.

\subsubsection{Optimization Objective}
Given a dataset $\mathcal{D}=\{(S_i,y_i)\}_i$, where $y_i\in\{0,1\}$ indicates whether the $i$-th text $S_i$ is human-written ($0$) or machine-generated ($1$), we introduce a latent retention mask $z_i\in\{0,1\}^{n_i}$ for the $n_i$ spans in $S_i$. 
Here, $z_{i,j}=1$ indicates that the $j$-th span is retained as machine-indicative evidence, while $z_{i,j}=0$ indicates that it is regarded as human-like and filtered out. 
Notably, $z_i$ does not represent the true authorship of each span. 
Instead, it represents a detection-oriented retention decision from a distributional perspective, since hidden human-like spans may also be generated by machines yet remain close to the human distribution. 
The retained part of $S_i$ is then defined as $\hat{S}_i=S_i\odot z_i$, where $\odot$ denotes element-wise masking.

The analysis in Section \ref{sec: theoretical_improvemenets} suggests that reducing the influence of hidden human-like spans can lower the sentence complexity of detection. 
Accordingly, a suitable optimization strategy is not to directly maximize the likelihood of the complete text $S_i$, but to optimize the detector based on the retained machine-indicative text $\hat{S}_i$. Let the detection model be parameterized by $\theta$. 
Since latent retention mask $z_i$ is unobserved, we maximize the marginal likelihood of the observed data by summing over all possible latent retention masks:

\begin{equation}
\label{eq: objective}
    \hat{\theta}=\arg\max_{\theta} \sum\nolimits_{(S_i,y_i)\in\mathcal{D}} \log \sum\nolimits_{z_i} p\left(y_i,S_i, z_i ; \theta\right).
\end{equation}

\subsubsection{Detector Training}
The objective in Eq. (\ref{eq: objective}) involves the latent retention mask $z_i$, which is unobserved. 
A natural way to optimize such a latent-variable objective is the Expectation-Maximization (EM) algorithm \cite{dempster1977maximum}. 
At iteration $t$, EM alternates between two steps.

In the E-step, it computes the expected complete-data log-likelihood under the current posterior of $z_i$:
$$Q(\theta;\theta^t)=\sum_{(S_i,y_i)}\sum_{z_i}p(z_i|S_i,y_i;\theta^t)\log p(y_i,S_i, z_i;\theta).$$ 

In the M-step, it updates the detector by maximizing this expectation:
$$\theta^{t+1}=\arg\max_{\theta}Q(\theta;\theta^t).$$

However, exact EM is computationally infeasible in our setting, because enumerating all possible span masks $z_i$ requires $\mathcal{O}(2^{n_i})$ computations for a text with $n_i$ spans. 
Therefore, we adopt a hard-EM-inspired approximation \cite{wenequal}, which replaces the full posterior expectation with a single most plausible latent mask. 
This leads to the following coordinate-ascent procedure:
\begin{itemize}[leftmargin = *,topsep=0cm]
    \item \textbf{Hard E-step}:  The E-step can be accomplished in a hard manner by choosing the best-fit $z_i$:
    \begin{equation}
        Q(\theta;\theta^t)=\sum\nolimits_{(S_i,y_i)}\log p(y_i,S_i,z_i;\theta),
        \nonumber
    \end{equation}
    \begin{equation}
        \text{ where } z_i=\arg\max_{z_i}p(z_i|S_i,y_i;\theta^t).
        \label{eq: hard-E}
    \end{equation}
    \item \textbf{Hard M-step}:  Maximize $Q(\theta,\theta^t)$ over $\theta$:
    \begin{equation}
        \theta^{t+1}=\arg\max_{\theta}Q(\theta;\theta^t).
        \label{eq: hard-M}
    \end{equation}
\end{itemize}
Compared with classical EM, hard EM avoids summing over all possible $z_i$ and therefore provides a more tractable optimization strategy. 
We also provide and evaluate a classical EM version in Section VI and Section VII-G of the Supplementary Material.

Given an estimated mask $z_i^t$, the hard M-step can be implemented by training the detector on the retained text $\hat{S}_i=S_i\odot z_i^t$. 
Specifically, the objective can be written as the standard binary log-likelihood:
\begin{equation}
    Q(\theta,\theta^t)=\sum_{(S_i,y_i)} y_i\log\left(\hat{y}_i\right)+(1-y_i)\log\left(1-\hat{y}_i\right).
    \label{eq: Q_equation}
\end{equation}
where $\hat{y}_i=f(S_i\odot z_i^t,\theta)$. 
Thus, the detector parameters can be updated through standard gradient-based optimization. 
The remaining key problem is how to obtain an operational estimate of $z_i^t$.

According to Eq. \ref{eq: hard-E}, the exact hard E-step requires the posterior $p(z_i|S_i,y_i;\theta^t)$. 
Although $y_i$ is available during supervised training, relying on $y_i$ to estimate $z_i$ would make the procedure inconsistent with inference, where the label is to be predicted. 
To this end, revisiting the posterior of the $z_i$:
\begin{equation}
    p\left(z_i \mid S_i, y_i ; \theta^{t}\right)=\frac{p\left(y_i \mid S_i, z_i ; \theta^{t}\right) p\left(z_i \mid S_i ; \theta^{t}\right)}{\sum_{z_i^{\prime}} p\left(y_i \mid S_i, z_i^{\prime} ; \theta^{t}\right) p\left(z_i^{\prime} \mid S_i ; \theta^{t}\right)}.
    \nonumber
\end{equation}
We have $z_i=\arg\max_{z_i}p\left(y_i \mid S_i, z_i ; \theta^{t}\right) p\left(z_i \mid S_i ; \theta^{t}\right)$, which requires $\mathcal{O}(2^n)$ computations. Considering that the effect of $z_i$ on $p\left(z_i \mid S_i ; \theta^{t}\right)$ is often much larger than the change in $p\left(y_i \mid S_i, z_i ; \theta^{t}\right)$, which allows us to approximate $z_i\approx\arg\max_{z_i}p\left(z_i \mid S_i ; \theta^{t}\right)$ \footnote{
The approximation is mild because the posterior term $p(z_i|S_i;\theta^t)$ usually dominates. Let 
$z_1=\arg\max_{z_i}p(z_i|S_i;\theta^t)$ and 
$z_2=\arg\max_{z_i}p(y_i|S_i,z_i;\theta^t)p(z_i|S_i;\theta^t)$. 
If they differ in $k$ positions, confident subsequence estimates, e.g., $>0.95$ or $<0.05$, yield 
$p(z_1|S_i;\theta^t)>20^k p(z_2|S_i;\theta^t)$. 
Therefore, choosing $z_2\neq z_1$ would require the paragraph-level likelihood term to offset an exponential gap, which is unlikely when only limited subsequences are modified.
}, reducing the complexity to $\mathcal{O}(n)$.
% (Section \eqref{sec: approximation} of the Supplementary Material details the rationale for this approximation).

In practice, we use the current detector to estimate the retention mask. 
Recall that $f(S,\theta)$ outputs the confidence that $S$ is machine-generated. 
Therefore, a span with a very low score is regarded as confidently human-like and should be filtered, while the remaining spans are retained as machine-indicative evidence. 
A naive retention rule is
\begin{equation}
    z_{i,j}=I\left(f(S_{i,j},\theta^t)\ge 0.5\right),
    \label{eq: origin_z}
\end{equation}
where $I(\cdot)$ is the indicator function. 
However, sentence-level estimates can be unstable, and directly using the standard threshold $0.5$ may incorrectly filter spans that still contain useful machine-indicative signals.

To mitigate the risk of mis-filtering, we introduce two practical designs. First, we extend sentence-level retention to sequence-level retention: each $S_{i,j}$ contains at most $k$ sentences, where $k=1$ degenerates to the sentence-level case. 
Second, we impose two constraints to prevent over-filtering:
\begin{itemize}[leftmargin=*]
    \item A stricter threshold $r_e$ is used to select only confidently human-like spans, where $r_e\ll 0.5$. For example, we use $r_e=0.01$ in our experiments.
    \item A maximum filtering ratio $\tau$ is used to limit the number of filtered spans. This avoids making a prediction based on too few remaining spans.
\end{itemize}

Based on these constraints, a span is filtered only when it is both confidently human-like and among the lowest-scored spans. 
Let $\mathcal{B}_{\tau n_i}(\{f(S_{i,j'},\theta^t)\}_{j'})$ denote the index set of the $\tau n_i$ smallest detector scores. 
Then the retention mask is computed as
\begin{equation}
    z_{i,j} =
\begin{cases}
0, & f(S_{i,j}, \theta^t) < r_e \ \text{and}\ j \in \mathcal{B}_{\tau n_i}(\{f(S_{i,j'},\theta^t)\}_{j'}),\\
1, & \text{otherwise}.
\end{cases}
\label{eq: mask}
\end{equation}
In this way, only a small number of highly confident human-like spans are removed, while most spans are preserved for detection. 
This conservative design follows the principle that it is preferable to leave some potential human-like noise in the text rather than mistakenly remove spans that contain valuable machine-indicative evidence.

\subsubsection{Detector Inference}

Under the hard-EM-inspired optimization, inference is performed by estimating the latent retention mask $z_i$ according to Eq. \ref{eq: mask}, and then detecting on the retained machine-indicative text:
\begin{equation}
    \hat{y}_i=f(S_i\odot z_i,\theta).
    \nonumber
\end{equation}
As shown in Fig. \ref{fig: framework}, the same detector is used twice: the first pass estimates which spans should be retained, and the second pass predicts the label based on the retained text. 
Therefore, the stacked inference procedure is not an independent heuristic, but the test-time realization of the latent-variable formulation and the conceptual filtering principle introduced above.

\subsubsection{Overall Framework}

Alg. \ref{alg: detection} summarizes the overall procedure of the proposed stacked enhancement framework.

\begin{itemize}[leftmargin=*]
    \item For \textbf{model inference} (Lines 1-6): 
    (1) the input text is first split into a set of smaller subsequences (Line 2); 
    (2) the current detector estimates the latent retention mask $z_i$ according to Eq. \ref{eq: mask}, where confidently human-like subsequences are filtered out, and the remaining subsequences are retained as machine-indicative evidence (Line 3); 
    (3) the retained subsequences are concatenated into the retained text $\hat{S}_i=S_i\odot z_i$ (Line 4); and 
    (4) the same detector $f(S,\theta)$ is applied again to $\hat{S}_i$ to produce the final prediction for the entire text (Line 5).

    \item For \textbf{model training} (Lines 7-15): 
    (1) in the hard E-step (Line 10), the current detector $f(S,\theta^t)$ estimates the best-fit latent retention masks $z_i$ for training samples. This yields the retained text $\hat{S}_i=S_i\odot z_i$ for each sample and allows us to compute $Q(\theta,\theta^t)$ in Eq. \ref{eq: Q_equation}; 
    (2) in the hard M-step (Line 11), the detector is updated on the retained texts, producing the next parameter estimate $\theta^{t+1}$.
\end{itemize}
\begin{algorithm}[t]
\caption{Stacked Detection Framework}
\label{alg: detection}
\begin{algorithmic}[1]
\Input Train data $\mathcal{D}=\{(S_i,y_i)\}_{i=1}^N$, the detection model $f(S,\theta^0)$, training epochs $T$, filtering ratio $\tau$, E-step detection threshold $r_e$, and learning rate $\eta$.
\Function{Inference}{$S_i,f(S,\theta^t)$}{\color{red}\Comment{Inference}}
    \State Split $S_i$ into a set of sequences $\{S_{i,j}\}$, where each $S_{i,j}$ contains at most $k$ sentences.
    \State Calculate $z_i$ according to Eq. (\ref{eq: mask}).
    \State $\hat{S}_i=S_{i}\odot z_i$.
    \State \Return $f(\hat{S}_i,\theta)$.
\EndFunction
\Function{Train}{$\mathcal{D}, f(S,\theta^0)$}{\color{red}\Comment{Training}}
    \For{$t=0$ {\bfseries to} $T-1$}
   \For{each batch of samples $\mathcal{D}_B\sim \mathcal{D}$}
   \State Calculate $\mathcal{Q}(\theta,\theta^t)$ according to Eq. (\ref{eq: Q_equation}), where $\hat{y}_i$ = Inference ($S_i, f(S,\theta^{t})$). {\color{blue}\Comment{E-step}}
   \State $\theta^{t+1}=\theta+\eta\nabla_\theta\mathcal{Q}(\theta,\theta^t)$. {\color{blue}\Comment{M-step}}
   \EndFor
   \EndFor
   \State \Return the trained model $f(S,\theta^T)$.
\EndFunction
\end{algorithmic}
\end{algorithm}
\subsection{Framework Analysis}

\subsubsection{The Validity under Imperfect Filtering}
Undeniably, even with stricter thresholds $r_e$ and maximum filter ratio $\tau$, the filtering component still risks excluding discriminative sub-text. To theoretically understand this risk, we further analyze an imperfect
filtering case as follows.
\begin{theorem}[\textbf{Validity under Imperfect Filtering}]
\label{theorem: imperfect}
Consider the MGT detection under the assumption of Theorem \ref{theorem: complexity_noniid}. If we filter an $\alpha_s$ ($<\alpha$) proportion of hidden human-like spans and mistakenly removes
an $\alpha_h$ proportion of machine-indicative spans, where
$0\leq \alpha_s<\alpha$, $0\leq \alpha_h<1-\alpha$, and
$\alpha_s+\alpha_h<1$. then the imperfectly filtered text has a larger TV lower bound than the original text in Theorem \ref{theorem: complexity_noniid} when
\[
1-\alpha-\alpha_h>
(1-\alpha)\sqrt{1-\alpha_s-\alpha_h},
\]
and
\[
\frac{1}{n}\sum_{j=1}^{L}(c_j-1)\rho_j
<
\frac{
\delta\left[
1-\alpha-\alpha_h
-
(1-\alpha)\sqrt{1-\alpha_s-\alpha_h}
\right]
}{
2\left(1-\sqrt{1-\alpha_s-\alpha_h}\right)
}.
\]
\end{theorem}
When $\alpha_s$ and $\alpha_h$ are small, we have
$
\sqrt{1-\alpha_s-\alpha_h}
\approx
1-\frac{\alpha_s+\alpha_h}{2}
$.
Thus, the first condition approximately becomes
$\alpha_s>
\frac{1+\alpha}{1-\alpha}\alpha_h$.
This condition is mild under conservative filtering: the filter correctly removes sufficiently more hidden human-like spans than mistakenly removed machine-indicative spans. Moreover, the second condition becomes
\[
\lambda_n
\lesssim
\delta\cdot
\frac{
(1-\alpha)\alpha_s-(1+\alpha)\alpha_h
}{
2(\alpha_s+\alpha_h)
}.
\]
When $\alpha_h\ll \alpha_s$, the right-hand side is close to
$(1-\alpha)\delta/2$, which corresponds to the standard effective-signal
condition in the non-IID setting (Theorem 1 in \cite{dhurandhar2013auto}). Therefore, conservative filtering is
theoretically justified: it is preferable to leave some hidden
human-like spans unfiltered rather than mistakenly remove too many
machine-indicative spans. The proof is provided in Section IV-E of the Supplementary Material.

% \subsubsection{The Validity of the Proposed Framework}

% Undeniably, even with stricter thresholds $r_e$ and maximum filter ratio $\tau$, the filtering component still risks excluding discriminative sub-text. Fortunately, our iterative stacked optimization architecture ensures its effectiveness through a self-reinforcing mechanism: (1) in the E-step, the detector, based on information bottleneck theory \cite{guan2019towards}, identifies and filters out high confidence "human" sequences, thereby supplying more distinctive remaining texts to the same detector in the M-step; (2) in the M-step, the detector learns on the remaining texts, leading to improvements that are used in the following E-step for even more accurate filtering of "human" sequences. This interwoven process leads to self-enhancement. A more formal validity description is given in Section \eqref{sec: framework_analysis} of the Supplementary Material.

\subsubsection{Time Complexity}
\label{sec: complexity}
For Transformer-based detectors, assuming the text length is $N$ and the embedding dimension is $d$, the time complexity of the original detector is $O(dN^2)$. In our stacked detection framework, the E-step divides the long text into several sequences of lengths $\{N_i\}_i$, resulting in a time complexity of $\mathcal{O}(\sum_idN_i^2)=\mathcal{O}(dN^2)$, which is usually lower than that of the original detector in practice. For the M-step, the complexity is also $\mathcal{O}(dN^2)$. Since the length of the filtered text does not exceed $N$, it is also not higher than that of the original detector. Therefore, our complexity is $\mathcal{O}(dN^2)$, and the actual running time does not exceed twice that of the original detector. We will further discuss the time complexity from the perspective of empirical experiments in Section \ref{sec: running_time}.

\section{Experiments}
\label{sec: experiment}

\subsection{Experiment Setup}
\subsubsection{Datasets}

The experiments are conducted on the MGT detection benchmark, MGTBench \cite{he2024mgtbench}, and we use four datasets as follows:

\begin{itemize}[leftmargin=*]
    \item \textbf{Essay} \cite{verma2024ghostbuster}. This dataset contains 1,000 essays collected from IvyPanda, covering diverse subjects and education levels. ChatGPT-turbo is first used to generate a content-aligned $<prompt>$ for each essay, and the prompt is then queried to multiple LLMs, including ChatGPT, GPT-4, ChatGPT-turbo, ChatGLM, Dolly, and Claude, to obtain the corresponding MGTs.
    
    \item \textbf{Reuters} \cite{verma2024ghostbuster}. This dataset is built from the Reuters 50-50 authorship identification dataset, which contains 1,000 news articles written by 50 journalists. Similar to the Essay dataset, ChatGPT-turbo is used to generate a $<headline>$ for each article, which is then used as the prompt for different LLMs, including ChatGPT, GPT-4, ChatGPT-turbo, ChatGLM, Dolly, and Claude.
    
    \item \textbf{SQuAD1} \cite{he2024mgtbench}. This dataset is derived from SQuAD1 \cite{rajpurkar2016squad} and contains 1,000 context-based questions. Each sample includes one human-written answer and multiple LLM-generated answers from ChatGPT, GPT-4, ChatGPT-turbo, ChatGLM, and StableLM. Following MGTBench, each sample is constructed by concatenating the question and answer, e.g., Q1+human answer or Q1+machine answer. The shared question part introduces common human-written context, allowing us to evaluate detection when human-like contextual spans are present.
    
    \item \textbf{DetectRL} \cite{wu2024detectrl}. This dataset contains texts from arXiv abstracts, XSum news, Writing Prompts stories, and Yelp reviews, with MGTs generated by GPT-3.5-turbo, PaLM-2-bison, Claude-instant, and Llama-2-70b. It also provides paraphrased texts generated by Dipper \cite{krishna2023paraphrasing}, LLM-based polishing, and back translation.
\end{itemize}
All datasets are randomly divided into the training, validation, and test sets with a ratio of 2: 1: 1.

% Table generated by Excel2LaTeX from sheet 'chatgpt'
\begin{table*}[t]
  \centering
  \caption{Performance concerning TPR@FPR-0.5\%. Detectors are trained on ChatGPT text.}
  \renewcommand\arraystretch{0.6}
  \begin{adjustbox}{width=1.\textwidth}
  \setlength{\tabcolsep}{0.02\linewidth}
    \begin{tabular}{c|c|cccccc|c}
    \hline\noalign{\smallskip} 
    Dataset & Method & ChatGPT  & GPT-4  & GPT-turbo & ChatGLM  & Dolly  & Claude/StableLM  & Avg. \\
    \noalign{\smallskip} \hline\noalign{\smallskip} 
        \multirow{13}[0]{*}{Essay} & Log-Likelihood & $24.08_{\pm22.67}$ & $37.70_{\pm30.63}$ & $23.12_{\pm24.38}$ & $5.86_{\pm7.02}$ & $12.45_{\pm12.04}$ & $2.48_{\pm3.06}$ & 17.62  \\
          % & Rank  & $55.60_{\pm3.65}$ & $51.31_{\pm5.91}$ & $65.84_{\pm6.93}$ & $53.49_{\pm6.09}$ & $35.11_{\pm4.81}$ & $25.12_{\pm5.07}$ & 47.75  \\
          & Log-Rank & $28.72_{\pm26.90}$ & $46.48_{\pm36.20}$ & $25.04_{\pm26.83}$ & $28.19_{\pm27.90}$ & $17.34_{\pm15.65}$ & $2.96_{\pm3.73}$ & 24.79  \\
          & DetectGPT & $37.04_{\pm10.21}$ & $24.75_{\pm10.03}$ & $5.52_{\pm1.93}$ & $21.45_{\pm14.25}$ & $15.45_{\pm7.05}$ & $4.48_{\pm2.57}$ & 18.12  \\
          & F-DetectGPT & $4.24_{\pm1.51}$ & $3.85_{\pm2.05}$ & $31.28_{\pm3.18}$ & $35.74_{\pm3.97}$ & $0.00_{\pm0.00}$ & $0.16_{\pm0.20}$ & 12.55  \\
          & RepreGuard &$68.00_{\pm11.09}$ & $62.62_{\pm24.10}$ & $0.32_{\pm0.16}$ & $31.89_{\pm2.22}$ & $20.86_{\pm3.22}$ & $0.16_{\pm0.20}$ & 30.64 \\
    & Lastde &$94.24_{\pm1.83}$ & $88.77_{\pm3.22}$ & $48.96_{\pm3.15}$ & $98.39_{\pm1.08}$ & $79.57_{\pm5.36}$ & $15.52_{\pm3.60}$ & 70.91 \\
    & Binoculars &$91.60_{\pm7.82}$ & $94.26_{\pm2.49}$ & $85.36_{\pm18.99}$ & $98.55_{\pm1.03}$ & $62.83_{\pm17.81}$ & $25.20_{\pm12.61}$ & 76.30 \\

          \noalign{\smallskip} \cline{2-9}\noalign{\smallskip} 
          & ChatGPT-D & $80.08_{\pm7.56}$ & $78.11_{\pm10.49}$ & $39.12_{\pm9.24}$ & $94.30_{\pm4.04}$ & $34.42_{\pm5.75}$ & $1.60_{\pm1.01}$ & 54.61  \\
          &\cellcolor{gray!20} \textbf{ChatGPT-STK} &\cellcolor{gray!20} $86.56_{\pm9.62}$ &\cellcolor{gray!20} $83.61_{\pm11.57}$ &\cellcolor{gray!20} $46.32_{\pm9.41}$ &\cellcolor{gray!20} $96.47_{\pm3.68}$ &\cellcolor{gray!20} $44.98_{\pm10.42}$ &\cellcolor{gray!20} $4.24_{\pm2.98}$ &\cellcolor{gray!20} 60.36  \\
          & OpenAI-D & $78.32_{\pm39.18}$ & $87.70_{\pm20.72}$ & $70.24_{\pm35.62}$ & $96.55_{\pm5.32}$ & $66.70_{\pm32.73}$ & $22.56_{\pm17.88}$ & 70.34  \\
          &\cellcolor{gray!20} \textbf{OpenAI-STK} &\cellcolor{gray!20} $96.96_{\pm1.83}$ &\cellcolor{gray!20} $\uline{96.89}_{\pm1.73}$ &\cellcolor{gray!20} $88.48_{\pm2.79}$ &\cellcolor{gray!20} $98.63_{\pm0.94}$ &\cellcolor{gray!20} $\textbf{81.55}_{\pm7.87}$ &\cellcolor{gray!20} $28.24_{\pm7.54}$ &\cellcolor{gray!20} 81.79  \\
          & MPU   & $\textbf{99.92}_{\pm0.16}$ & $\textbf{99.26}_{\pm0.40}$ & $67.92_{\pm12.35}$ & $\textbf{99.60}_{\pm0.51}$ & $76.14_{\pm6.98}$ & $59.92_{\pm18.84}$ & 83.79  \\
          &\cellcolor{gray!20} \textbf{MPU-STK} &\cellcolor{gray!20} $\textbf{99.92}_{\pm0.16}$ &\cellcolor{gray!20} $\textbf{99.26}_{\pm0.40}$ &\cellcolor{gray!20} $65.44_{\pm16.47}$ &\cellcolor{gray!20} $\textbf{99.60}_{\pm0.51}$ &\cellcolor{gray!20} $\uline{79.06}_{\pm6.65}$ &\cellcolor{gray!20} $\textbf{78.64}_{\pm19.68}$ &\cellcolor{gray!20} \uline{86.99}  \\
          & RADAR & $96.88_{\pm2.25}$ & $96.56_{\pm0.56}$ & $\uline{92.64}_{\pm4.47}$ & $98.96_{\pm0.90}$ & $65.75_{\pm8.58}$ & $58.64_{\pm8.36}$ & 84.90  \\
          &\cellcolor{gray!20} \textbf{RADAR-STK} &\cellcolor{gray!20} $\uline{98.16}_{\pm1.38}$ &\cellcolor{gray!20} $95.82_{\pm1.78}$ &\cellcolor{gray!20} $\textbf{94.64}_{\pm3.89}$ &\cellcolor{gray!20} $\uline{99.12}_{\pm0.59}$ &\cellcolor{gray!20} $70.39_{\pm8.22}$ &\cellcolor{gray!20} $\uline{64.96}_{\pm7.80}$ &\cellcolor{gray!20} \textbf{87.18}  \\
          \noalign{\smallskip} \hline\hline\noalign{\smallskip} 
    \multirow{13}[0]{*}{Reuters} & Log-Likelihood & $77.84_{\pm5.19}$ & $14.88_{\pm5.98}$ & $86.08_{\pm3.38}$ & $93.76_{\pm2.03}$ & $11.20_{\pm4.45}$ & $15.04_{\pm6.86}$ & 49.80  \\
          % & Rank  & $48.88_{\pm1.59}$ & $35.92_{\pm2.88}$ & $58.40_{\pm3.94}$ & $40.56_{\pm1.85}$ & $18.56_{\pm2.27}$ & $6.24_{\pm1.87}$ & 34.76  \\
          & Log-Rank & $82.40_{\pm5.24}$ & $25.92_{\pm7.08}$ & $90.96_{\pm4.12}$ & $96.80_{\pm0.88}$ & $14.00_{\pm4.82}$ & $17.60_{\pm8.29}$ & 54.61  \\
          & DetectGPT & $4.40_{\pm2.62}$ & $0.64_{\pm0.54}$ & $2.32_{\pm1.87}$ & $2.56_{\pm2.80}$ & $0.48_{\pm0.47}$ & $3.04_{\pm1.61}$ & 2.24  \\
          & F-DetectGPT & $48.00_{\pm9.48}$ & $6.80_{\pm1.88}$ & $92.96_{\pm1.65}$ & $88.96_{\pm4.80}$ & $0.00_{\pm0.00}$ & $0.48_{\pm0.39}$ & 39.53  \\
          & RepreGuard &$26.88_{\pm19.31}$ & $9.04_{\pm6.02}$ & $0.08_{\pm0.16}$ & $0.80_{\pm1.01}$ & $0.00_{\pm0.00}$ & $0.00_{\pm0.00}$ & 6.13 \\
    & Lastde &$92.64_{\pm1.06}$ & $88.96_{\pm1.71}$ & $87.36_{\pm3.51}$ & $98.56_{\pm0.65}$ & $66.00_{\pm3.27}$ & $9.68_{\pm3.04}$ & 73.87 \\
    & Binoculars &$95.84_{\pm0.93}$ & $70.88_{\pm3.93}$ & $96.08_{\pm0.93}$ & $99.28_{\pm0.30}$ & $40.88_{\pm4.94}$ & $26.16_{\pm4.18}$ & 71.52 \\

          \noalign{\smallskip} \cline{2-9}\noalign{\smallskip} 
          & ChatGPT-D & $98.00_{\pm2.25}$ & $94.32_{\pm3.97}$ & $96.08_{\pm2.23}$ & $98.48_{\pm0.78}$ & $59.76_{\pm13.36}$ & $11.84_{\pm6.11}$ & 76.41  \\
          &\cellcolor{gray!20} \textbf{ChatGPT-STK} &\cellcolor{gray!20} $99.28_{\pm0.39}$ &\cellcolor{gray!20} $96.16_{\pm1.15}$ &\cellcolor{gray!20} $98.08_{\pm1.17}$ &\cellcolor{gray!20} $98.72_{\pm0.47}$ &\cellcolor{gray!20} $64.56_{\pm8.32}$ &\cellcolor{gray!20} $30.32_{\pm8.23}$ &\cellcolor{gray!20} 81.19  \\
          & OpenAI-D & $96.88_{\pm4.26}$ & $84.08_{\pm9.42}$ & $96.56_{\pm5.32}$ & $98.00_{\pm1.13}$ & $49.44_{\pm5.83}$ & $19.92_{\pm5.21}$ & 74.15  \\
          &\cellcolor{gray!20} \textbf{OpenAI-STK} &\cellcolor{gray!20} $99.52_{\pm0.30}$ &\cellcolor{gray!20} $95.36_{\pm2.19}$ &\cellcolor{gray!20} $99.76_{\pm0.20}$ &\cellcolor{gray!20} $98.48_{\pm0.53}$ &\cellcolor{gray!20} $62.72_{\pm5.44}$ &\cellcolor{gray!20} $39.44_{\pm7.20}$ &\cellcolor{gray!20} 82.55  \\
          & MPU   & $\textbf{100.00}_{\pm0.00}$ & $97.92_{\pm1.06}$ & $\uline{99.92}_{\pm0.16}$ & $\uline{99.60}_{\pm0.25}$ & $72.64_{\pm7.02}$ & $75.68_{\pm12.92}$ & 90.96  \\
          &\cellcolor{gray!20} \textbf{MPU-STK} &\cellcolor{gray!20} $\textbf{100.00}_{\pm0.00}$ &\cellcolor{gray!20} $\uline{98.08}_{\pm1.14}$ &\cellcolor{gray!20} $\textbf{100.00}_{\pm0.00}$ &\cellcolor{gray!20} $99.44_{\pm0.41}$ &\cellcolor{gray!20} $72.80_{\pm8.95}$ &\cellcolor{gray!20} $\uline{84.40}_{\pm10.39}$ &\cellcolor{gray!20} \uline{92.45}  \\
          & RADAR & $\textbf{100.00}_{\pm0.00}$ & $\textbf{99.92}_{\pm0.16}$ & $99.68_{\pm0.30}$ & $\textbf{99.92}_{\pm0.16}$ & $\textbf{89.68}_{\pm2.08}$ & $\textbf{95.68}_{\pm1.53}$ & \textbf{97.48}  \\
          &\cellcolor{gray!20} \textbf{RADAR-STK} &\cellcolor{gray!20} $\textbf{100.00}_{\pm0.00}$ &\cellcolor{gray!20} $\textbf{99.92}_{\pm0.16}$ &\cellcolor{gray!20} $99.68_{\pm0.30}$ &\cellcolor{gray!20} $\textbf{99.92}_{\pm0.16}$ &\cellcolor{gray!20} $\uline{89.68}_{\pm2.25}$ &\cellcolor{gray!20} $\textbf{95.68}_{\pm1.53}$ &\cellcolor{gray!20} \textbf{97.48}  \\
          \noalign{\smallskip} \hline
    \end{tabular}%
    \end{adjustbox}
  \label{tab: black_tpr_chatgpt}%
  \vspace{-0.3cm}
\end{table*}%

\subsubsection{Baselines}
We compare the proposed method with the following methods:

\begin{itemize}[leftmargin=*]
    \item \textbf{Log-Likelihood} \cite{solaiman2019release}. A zero-shot method that detects MGTs using the total log probability from gpt2-medium.
    \item \textbf{Log-Rank} \cite{gehrmann2019gltr}. A zero-shot method based on the average token log-rank computed by gpt2-medium.
    \item \textbf{DetectGPT} \cite{mitchell2023detectgpt}. A perturbation-based method that detects MGTs by measuring changes in log probability after text perturbation.
    \item \textbf{Fast-DetectGPT} \cite{bao2024fast}. An efficient variant of DetectGPT that replaces perturbation with a faster sampling-based strategy.
    \item \textbf{RepreGuard} \cite{chen2025repreguard}. A representation-based method that uses a proxy model to extract neural features and compute projection scores.
    \item \textbf{Lastde} \cite{xutraining}. A time-series-based method that combines local token-probability dynamics with global statistical features.
    \item \textbf{Binoculars} \cite{hans2024spotting}. A training-free method that compares scores from two closely related pre-trained language models.
    \item \textbf{ChatGPT-D} \cite{guo2023close}. A RoBERTa-based detector trained on HC3 using answer-only texts or QA pairs.
    \item \textbf{OpenAI-D} \cite{solaiman2019release}. A RoBERTa-based detector fine-tuned on GPT-2-generated texts.
    \item \textbf{MPU} \cite{tianmultiscale}. A RoBERTa-based detector trained with a multiscale positive-unlabeled learning framework.
    \item \textbf{RADAR} \cite{hu2023radar}. A robust detector trained adversarially with a paraphraser and a detector.
\end{itemize}

\subsubsection{Evaluation Metrics}
We first use the area under the receiver operating characteristic curve (AUROC). Besides, considering that a low false positive rate (i.e., human-generated texts being misclassified as machines) can mitigate repercussions for users \cite{fraser2025detecting}, we report performance as the true positive rate at a fixed false positive rate $K$ (TPR@FPR-$K$). In the experiments, $K$ is set to 0.5\%. To compare with GPTZero, which outputs hard labels, we also report the performance concerning Accuracy in Section VII-E of the Supplementary Material. All experiments are repeated 5 times. The best results are bolded, and the second-best results are underlined. 

\begin{figure*}[t]
	\centering
	\includegraphics[width=0.93\linewidth]{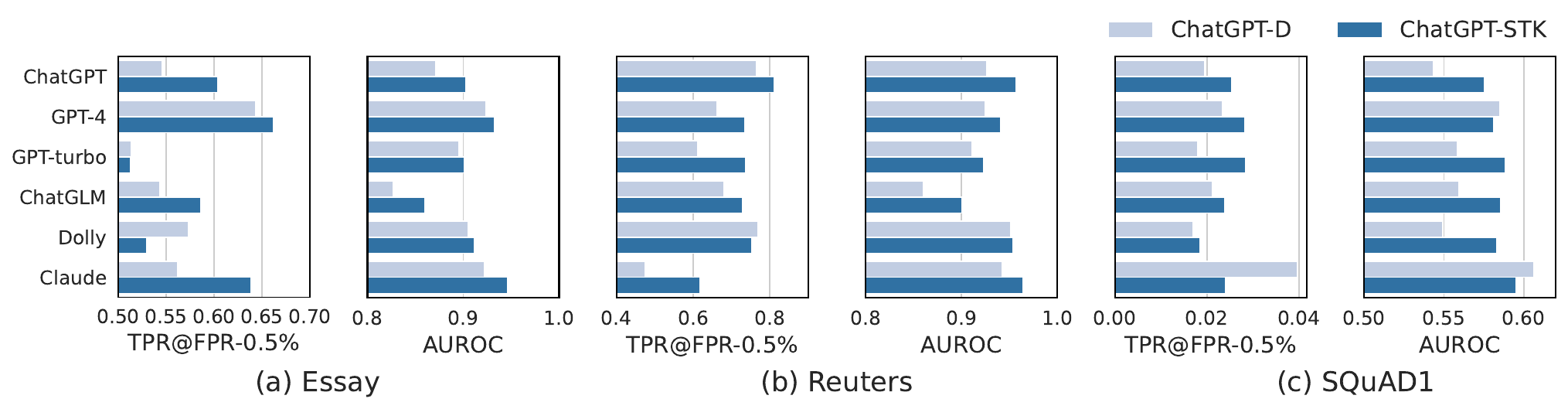}
 \vspace{-0.2cm}
	\caption{Average detection performance (x-axis) of detectors (ChatGPT-D and our boosting strategy ChatGPT-STK) tested across various LLMs, where these detectors are trained on texts generated by specific LLM (y-axis).}
	\label{fig: enhance_chatgpt}
    \vspace{-0.2cm}
\end{figure*}

\begin{figure*}[t]
	\centering
	\includegraphics[width=0.93\linewidth]{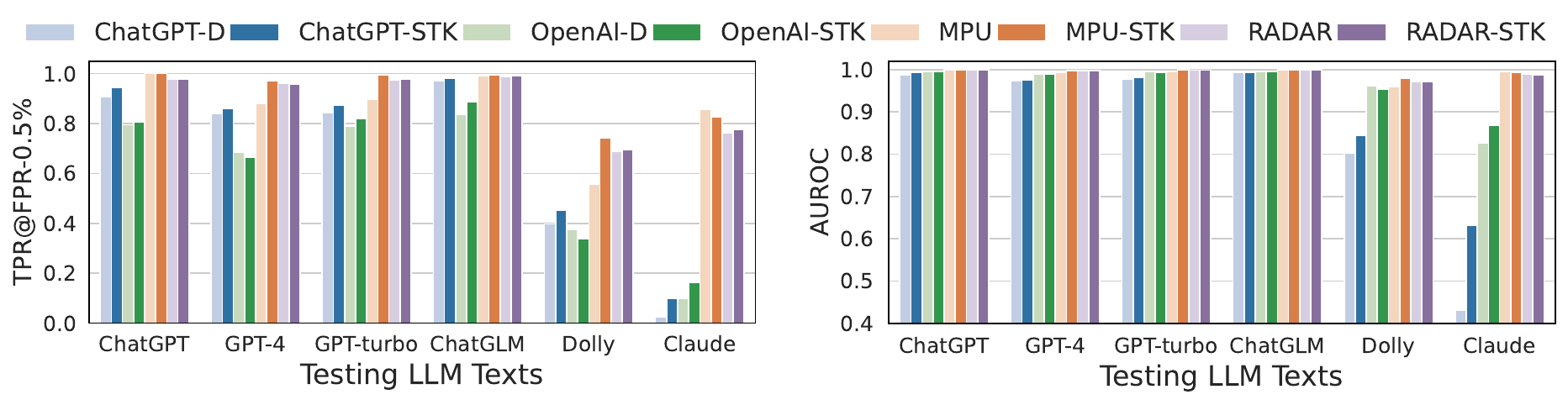}
 \vspace{-0.2cm}
	\caption{Performance under cross-domain setting. The Essay dataset served as the source domain, and the Reuters dataset as the target domain. The detector is trained on ChatGPT texts.}
	\label{fig: cross}
    \vspace{-0.2cm}
\end{figure*}

\subsection{Cross-LLM Performance}
\label{sec: cross-llm}
In real-world scenarios, detectors usually have no prior knowledge of the source LLM. Therefore, we train detectors on texts generated by a specific LLM and test them on texts generated by various LLMs. Table \ref{tab: black_tpr_chatgpt} reports the TPR@FPR-0.5\% when detectors are trained on ChatGPT texts and tested on different LLM texts. More results are provided in the Supplementary Material (Table I and Table II of Section VII-B).

First, the proposed stacked framework consistently improves the original detectors. For example, on the Essay dataset, ChatGPT-STK increases the average TPR@FPR-0.5\% of ChatGPT-D from 54.61\% to 60.36\%. Besides the average improvement, the proposed framework also shows enhancement potential in most specific LLM-testing cases. This indicates that reducing the influence of hidden human-like spans can improve detection even when the testing LLM differs from the training LLM, which is important for practical deployment. Moreover, cross-LLM performance is not always inferior to intra-LLM performance. This may depend on the generation quality and stylistic properties of different LLMs. For example, ChatGLM texts under the cross-LLM setting can be easier to detect than ChatGPT texts under the intra-LLM setting. Similar observations have also been reported in related work \cite{he2024mgtbench}. Finally, consistent with existing findings \cite{wu2024detectrl}, feature-based methods are generally less effective than model-based methods, suggesting that manually designed features may not sufficiently capture the complex and hidden human-like characteristics of MGTs.

Beyond training on ChatGPT-generated texts, we further evaluate detectors trained on texts from other LLMs. Fig. \ref{fig: enhance_chatgpt} presents the average performance of ChatGPT-D and ChatGPT-STK trained across various LLMs, where the y-axis denotes the LLM used for training and the x-axis denotes the average testing performance. Results for OpenAI-STK, MPU-STK, and RADAR-STK are shown in Fig. 1-3 of the Supplementary Material, and detailed results are reported in Tables III-XII of the Supplementary Material. The proposed framework consistently improves cross-LLM detection capability for these detectors.

\subsection{Cross-domain Performance}

In addition to cross-LLM generalization, we evaluate cross-domain performance, with the results shown in Fig. \ref{fig: cross}. In this setting, the Essay dataset is used as the source domain and the Reuters dataset as the target domain. This setting is challenging because the two domains differ substantially in writing style, topic distribution, and discourse structure.

The results show that detectors equipped with the proposed stacked framework achieve better performance in most cases. This indicates that the proposed framework is not limited to alleviating generator-specific distribution shifts, but can also improve robustness under domain shifts. One possible reason is that hidden human-like spans are often entangled with domain-dependent expressions, such as common phrases, templates, or stylistic patterns. By conservatively filtering confidently human-like subsequences and retaining more machine-indicative evidence, the proposed framework reduces the detector's reliance on domain-specific superficial cues. Consequently, the detector can make predictions based on more transferable signals, leading to improved cross-domain generalization.

\begin{figure}[t]
	\centering
	\includegraphics[width=0.93\linewidth]{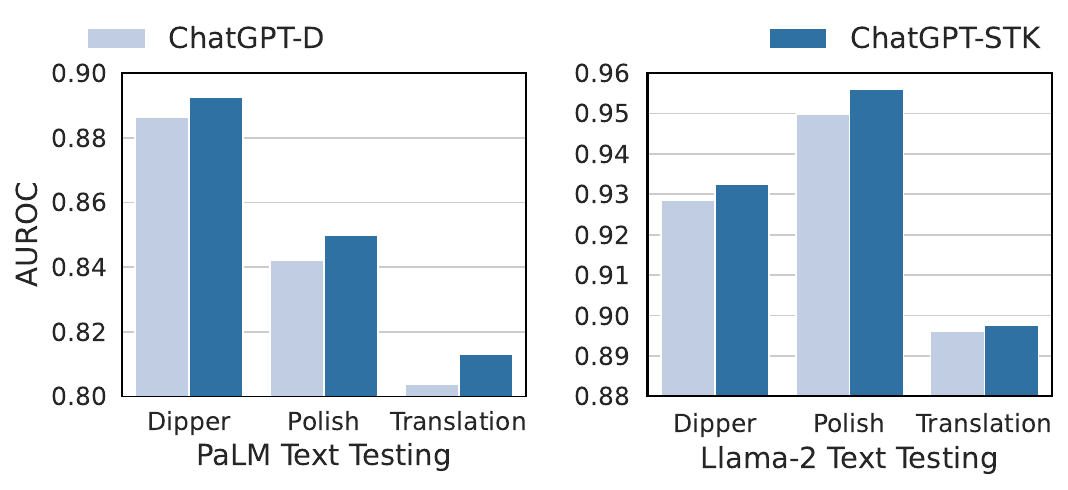}
 \vspace{-0.1cm}
	\caption{Enhancing the robustness of ChatGPT-D. Here we use three attacks: Dipper, Polish, and Translation.}
	\label{fig: attack}
    \vspace{-0.2cm}
\end{figure}

\begin{table*}[htbp]
  \centering
  \caption{Enhancements to the AUROC metric for sentence-based detector. Detector are trained on ChatGPT text.}
    \begin{tabular}{c|c|cccccc|c}
    \hline\noalign{\smallskip} 
    \multicolumn{1}{c|}{Dataset} & Method & \multicolumn{1}{c}{ChatGPT} & \multicolumn{1}{c}{GPT4All} & \multicolumn{1}{c}{ChatGPT-turbo} & \multicolumn{1}{c}{ChatGLM} & \multicolumn{1}{c}{Dolly} & \multicolumn{1}{c}{Claude} & \multicolumn{1}{|c}{Avg.} \\
    \noalign{\smallskip} \hline\hline\noalign{\smallskip} 
    \multicolumn{1}{c|}{\multirow{2}[0]{*}{Essay}} & SeqXGPT & \textbf{100} & 99.99 & 52.18 & 99.95 & 99.91 & 54.67 & 84.45 \\
          & \textbf{SeqXGPT-STK} & \textbf{100} & \textbf{100} & \textbf{53.17} & \textbf{100} & \textbf{99.99} & \textbf{58.75} & \textbf{85.32} \\
          \noalign{\smallskip} \hline\hline\noalign{\smallskip}
    \multicolumn{1}{c|}{\multirow{2}[0]{*}{Reuters}} & SeqXGPT & 99.34 & \textbf{99.48} & 61    & 99.54 & \textbf{99.51} & 60.78 & 86.61 \\
          & \textbf{SeqXGPT-STK} & \textbf{99.44} & 99.25 & \textbf{65.69} & \textbf{99.75} & 99.3  & \textbf{76.85} & \textbf{90.05} \\
           \noalign{\smallskip} \hline
    \end{tabular}%
  \label{tab: seqxgpt}%
\end{table*}%

\begin{table*}[t]
  \centering
  \caption{Detection performance (AUROC) on Essay dataset when the text length is at most 64. Detectors are trained on ChatGPT texts.}
  % \begin{adjustbox}{width=0.8\textwidth}
  % \setlength{\tabcolsep}{0.008\linewidth}
    \begin{tabular}{c|cccccc|c}
    \hline\noalign{\smallskip} 
    Method & ChatGPT & GPT4All & ChatGPT-turbo & ChatGLM & Dolly & Claude & Avg. \\
    \noalign{\smallskip} \hline\noalign{\smallskip} 
    ChatGPT-D & $89.94_{\pm1.13}$ & $86.29_{\pm2.93}$ & $85.13_{\pm2.60}$ & $96.66_{\pm0.99}$ & $70.23_{\pm2.41}$ & $64.79_{\pm3.25}$ & 82.17  \\
    \rowcolor{gray!20}\textbf{ChatGPT-STK} & $91.33_{\pm1.01}$ & $88.17_{\pm3.78}$ & $87.69_{\pm0.53}$ & $97.15_{\pm0.59}$ & $75.16_{\pm2.71}$ & $69.44_{\pm5.13}$ & 84.82  \\
    OpenAI-D & $93.50_{\pm1.22}$ & $93.10_{\pm1.94}$ & $89.93_{\pm1.52}$ & $98.09_{\pm0.33}$ & $85.36_{\pm1.86}$ & $72.74_{\pm2.85}$ & 88.79  \\
    \rowcolor{gray!20}\textbf{OpenAI-STK} & $95.68_{\pm0.49}$ & $94.21_{\pm1.13}$ & $93.69_{\pm0.80}$ & $98.59_{\pm0.40}$ & $88.84_{\pm0.60}$ & $81.67_{\pm2.05}$ & 92.11  \\
    MPU   & $96.28_{\pm0.42}$ & $95.77_{\pm0.37}$ & $96.01_{\pm0.81}$ & $99.24_{\pm0.14}$ & $81.00_{\pm1.07}$ & $85.57_{\pm0.67}$ & 92.32  \\
    \rowcolor{gray!20}\textbf{MPU-STK} & $96.13_{\pm0.27}$ & $95.66_{\pm0.40}$ & $95.81_{\pm0.80}$ & $99.29_{\pm0.12}$ & $83.38_{\pm1.43}$ & $86.11_{\pm0.57}$ & 92.72  \\
    RADAR & $98.63_{\pm0.17}$ & $96.25_{\pm0.81}$ & $97.87_{\pm0.40}$ & $99.30_{\pm0.11}$ & $92.29_{\pm0.95}$ & $93.31_{\pm0.84}$ & 96.28  \\
    \rowcolor{gray!20}\textbf{RADAR-STK} & $98.77_{\pm0.17}$ & $96.39_{\pm0.64}$ & $98.07_{\pm0.44}$ & $99.38_{\pm0.05}$ & $92.58_{\pm1.02}$ & $93.63_{\pm1.10}$ & 96.47  \\
    \noalign{\smallskip} \hline
    \end{tabular}%
    % \end{adjustbox}
  \label{tab: length64}%
\end{table*}%

\subsection{Robust Performance against Paraphrasing Attack}

Existing research \cite{sadasivan2023can} shows that MGT detectors are vulnerable to paraphrasing attacks, where attackers rewrite generated texts while preserving their semantics to evade detection. This setting is particularly challenging because paraphrasing can make MGTs more fluent and human-like, thereby weakening the discriminative evidence available to detectors.

To evaluate robustness under this scenario, we use three types of paraphrased texts from the DetectRL dataset, including Dipper, Polish, and Back Translation. The results are shown in Fig. \ref{fig: attack}, and more robustness results are provided in Section VII-D of the Supplementary Material. We observe that the proposed stacked framework improves the robustness of the original detector under these attacks. This suggests that, even when MGTs become more human-like after paraphrasing, filtering confidently human-like subsequences can still help the detector rely more on the remaining machine-indicative evidence, demonstrating the framework's practical value in adversarial detection scenarios.

\subsection{Enhancement of Auto-regressive-based Detector}

\begin{figure}[t]
	\centering
	\includegraphics[width=0.93\linewidth]{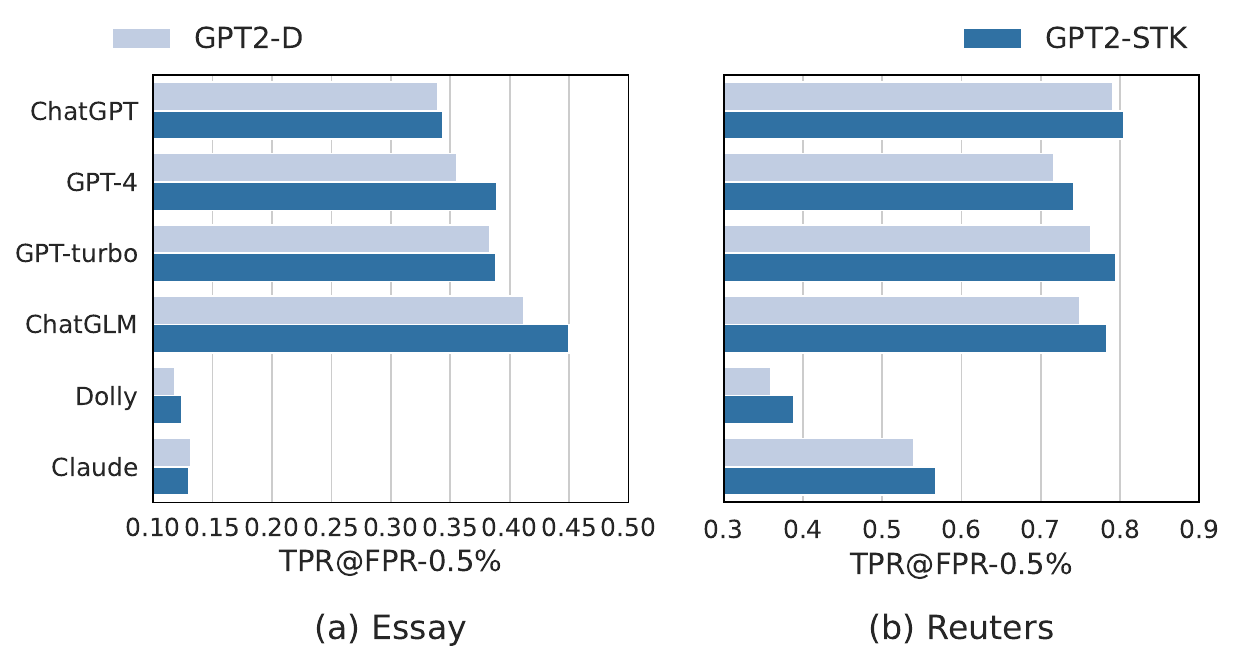}
 \vspace{-0.1cm}
	\caption{Enhancements to GPT2-based Detector, which is trained on ChatGPT texts.}
	\label{fig: enhance_gpt2}
    \vspace{-0.2cm}
\end{figure}

The proposed framework may affect auto-regressive-based detectors because filtering subsequences can disrupt local left-to-right contextual dependencies. To examine this issue, we use GPT-2 as the detection model, where the last-token embedding is fed into a fully connected layer for binary classification. The results are shown in Fig. \ref{fig: enhance_gpt2}. We observe that the proposed framework still improves the performance of the GPT-2-based detector.

This suggests that reducing the influence of hidden human-like spans can outweigh the potential loss caused by local context disruption. One possible reason is that important machine-indicative tokens can still exert persistent influence through the attention mechanism, allowing the detector to preserve useful discriminative signals after filtering. More theoretical discussion is provided in the second point of Section V-C of the Supplementary Material.

\subsection{Enhancements to Sentence-based Detector}
\label{sec: seqxgpt}

We further apply the proposed framework to the sentence-based detector SeqXGPT \cite{wang2023seqxgpt}, and the results are shown in Table \ref{tab: seqxgpt}. The proposed framework improves SeqXGPT's detection performance, demonstrating its flexibility beyond paragraph-level detectors. This suggests that the proposed latent filtering mechanism can also enhance detectors that rely on sentence-level predictions.
Although SeqXGPT outperforms feature-based methods in some settings, it is still less competitive than stronger model-based detectors. One possible reason is that directly aggregating sentence labels may be suboptimal for paragraph-level detection. When an MGT contains hidden human-like spans, these local spans may be predicted as human-like and bias the aggregated decision toward the human-written class. By filtering confidently human-like subsequences and retaining more machine-indicative evidence, the proposed framework alleviates this issue and improves the final text-level prediction.

\subsection{Performance Comparison under Shorter Texts}

Our work mainly focuses on paragraph-level detection, where relatively sufficient contextual information is available for identifying hidden human-like spans. For shorter texts, MGT detection is more challenging because the number of available sentences or spans is limited, and filtering may further reduce the evidence used for prediction. Nevertheless, since the proposed framework does not impose a strict requirement on text length, we further evaluate its effectiveness under shorter-text settings.

To this end, we test on the Essay dataset by limiting the maximum text length to 64 words. The results are shown in Table \ref{tab: length64}. Compared with the performance on original texts, detecting shorter texts is indeed more difficult, consistent with our theoretical findings that fewer sentences increase detection difficulty (Theorems \ref{theorem: complexity_noniid}). Even in this constrained setting, the proposed framework still provides consistent improvements, suggesting that conservative filtering can reduce hidden human-like spans while preserving useful machine-indicative evidence.

\subsection{Performance Comparison under Different Mixed Degrees}
\label{sec: mixed_degree}

\begin{figure*}[t]
	\centering
	\includegraphics[width=1.\linewidth]{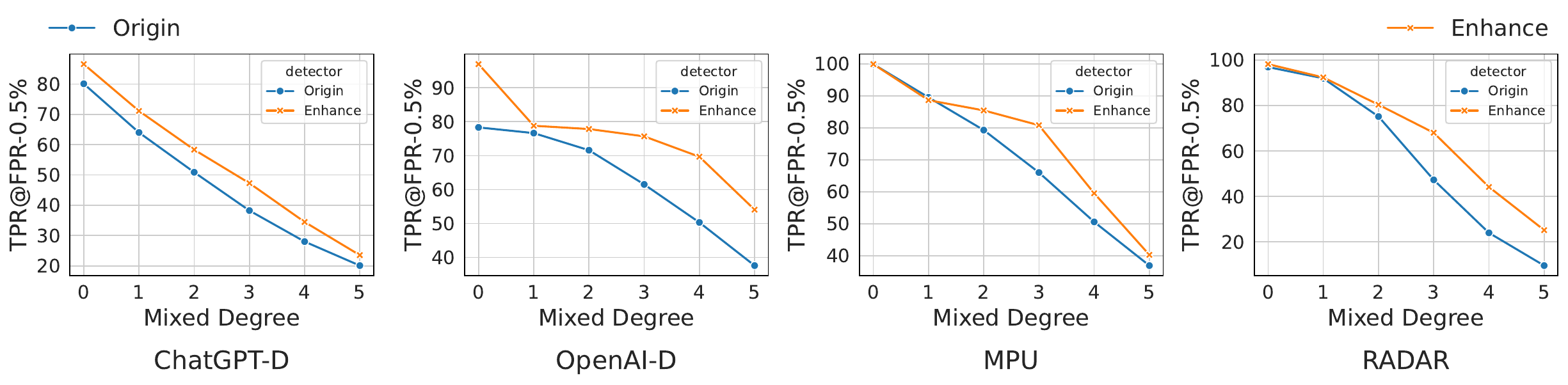}
  \vspace{-0.6cm}
	\caption{Performance concerning TPR@FPR-5\% at different mixing levels. These detectors are trained on ChatGPT texts.}
	\label{fig: mixed_degree}
\end{figure*}

% Table generated by Excel2LaTeX from sheet 'temp1'
\begin{table*}[t]
  \centering
  \caption{Performance (TPR@FPR-0.5\%) under different text-dropping strategies. The detectors are trained on ChatGPT texts.}
    \begin{tabular}{c|c|cccccc|c}
     \hline\noalign{\smallskip}
    Dataset & Method & ChatGPT & GPT4All & ChatGPT-turbo & ChatGLM & Dolly & Claude & Avg. \\
    \noalign{\smallskip}\hline\noalign{\smallskip}
    \multirow{3}[0]{*}{Essay} & ChatGPT-D & 80.08 & 78.11 & 39.12 & 94.3  & 34.42 & 1.6   & 54.61 \\
          & ChatGPT-rand & 76.56 & 78.28 & 25.2  & 84.18 & 37.77 & 2.56  & 50.76 \\
          & \textbf{ChatGPT-STK} & \textbf{86.56} & \textbf{83.61} & \textbf{46.32} & \textbf{96.47} & \textbf{44.98} & \textbf{4.24} & \textbf{60.36} \\
          \noalign{\smallskip}\hline\hline\noalign{\smallskip}
    \multirow{3}[0]{*}{Reuters} & ChatGPT-D & 98    & 94.32 & 96.08 & 98.48 & 59.76 & 11.84 & 76.41 \\
          & ChatGPT-rand & 95.52 & 89.92 & 89.28 & 95.52 & 58.32 & 18.64 & 74.53 \\
          & \textbf{ChatGPT-STK} & \textbf{99.28} & \textbf{96.16} & \textbf{98.08} & \textbf{98.72} & \textbf{64.56} & \textbf{30.32} & \textbf{81.19} \\
          \noalign{\smallskip}\hline
    \end{tabular}%
  \label{tab: random_filter}%
\end{table*}%

\begin{figure*}[t]
	\centering
	\includegraphics[width=0.93\linewidth]{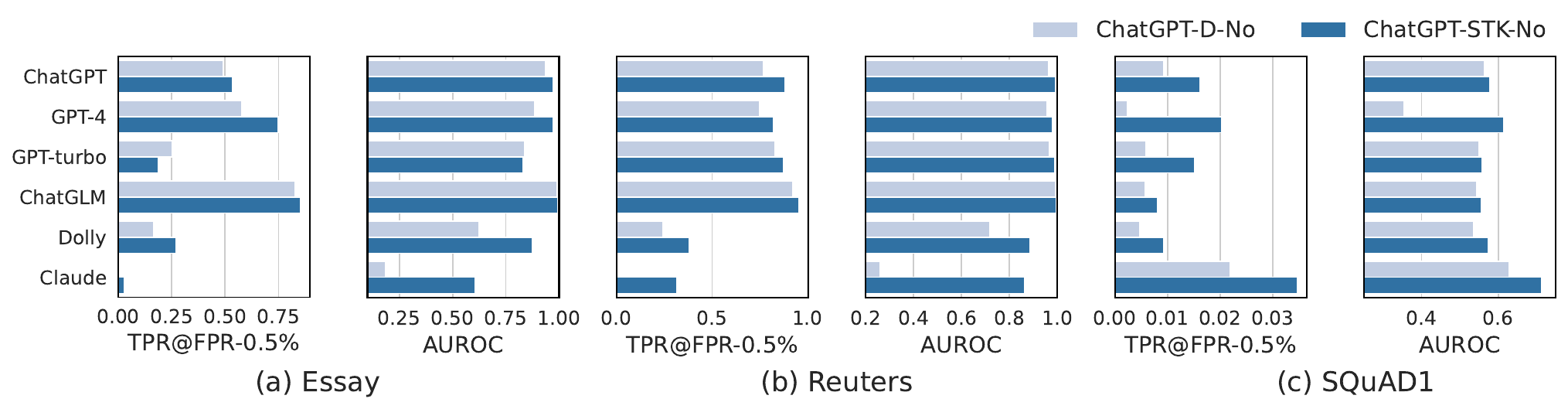}
 \vspace{-0.2cm}
	\caption{Performance (x-axis) of the un-fine-tuned detectors tested on various LLM texts (y-axis).}
	\label{fig: enhance_notrain}
    \vspace{-0.2cm}
\end{figure*}

We construct controlled test texts with different proportions of human-like spans based on the Essay dataset. Specifically, for each test text, we replace $n$ random sentences ($n$ ranges from 1 to 5) with human-written sentences under the same prompt, thereby simulating increasing proportions of human-like spans within MGTs. The results are shown in Table \ref{fig: mixed_degree}. First, as the proportion increases, all detectors gradually degrade, supporting our theoretical finding that human-like spans hinder detection. Second, the proposed framework shows larger advantages under higher proportions, further validating its effectiveness in reducing the influence of hidden human-like spans.

\begin{table*}[htbp]
\footnotesize
  \centering
  \caption{Running time comparison.}
  \begin{adjustbox}{width=0.75\textwidth}
  \setlength{\tabcolsep}{0.022\linewidth}{
    \begin{tabular}{c|ccc|ccc}
    \hline\noalign{\smallskip}
    \multirow{2}[0]{*}{Method} & \multicolumn{3}{c|}{Training Time (s)} & \multicolumn{3}{c}{Inference Time (s)} \\
    \noalign{\smallskip}\cline{2-7}\noalign{\smallskip}
          & Essay & Reuters & SQuAD1 & Essay & Reuters & SQuAD1 \\
          \noalign{\smallskip}\hline\noalign{\smallskip}
    Log-Likelihood & 42.17  & 56.29  & 20.43  &19.82  & 22.34  & 11.39  \\
    Log-Rank  & 72.04  & 81.32  & 31.69  & 35.13  & 39.32  & 17.98  \\
    DetectGPT & 3829.10  & 4356.37  & 442.74  & 1960.20  & 2269.46  & 858.28  \\
    F-DetectGPT & 51.39  & 66.77  & 20.10  & 113.69  & 126.78  & 39.60  \\
    RepreGuard & 274.65  & 297.58  & 234.77  & 104.14  & 123.41  & 68.55  \\
    Lastde & 168.67  & 184.79  & 96.78  & 80.53  & 89.79  & 51.64  \\
    Binoculars & 432.93  & 445.81  & 252.07  & 214.35  & 221.72  & 147.68  \\
    ChatGPT-D & 151.72  & 167.33  & 118.65  & 5.79  & 5.91  & 2.41  \\
    OpenAI-D & 156.23  & 169.09  & 129.37  & 5.81  & 5.99  & 2.57  \\
    MPU   & 172.83  & 189.46  & 141.20  & 5.83  & 5.87  & 2.62  \\
    RADAR & 326.79  & 331.63  & 254.16  & 26.54  & 27.03  & 22.49  \\
    \noalign{\smallskip}\hline\noalign{\smallskip}
    \textbf{ChatGPT-STK} & 269.85  & 252.18  & 139.54  & 10.75  & 10.89  & 5.67  \\
    \textbf{OpenAI-STK} & 286.79  & 255.36  & 148.24  & 11.17  & 10.92  & 6.08  \\
    \textbf{MPU-STK} & 301.51  & 289.77  & 151.92  & 11.26  & 11.69  & 6.37  \\
    \textbf{RADAR-STK} & 587.16  & 590.11  & 530.20  & 45.38  & 46.81  & 40.56  \\
    \noalign{\smallskip}\hline
    \end{tabular}%
    }
    \end{adjustbox}
  \label{tab: running_time}%
\end{table*}%

\subsection{Performance Comparison with Random Filtering}

To further evaluate whether the improvement comes from the proposed latent filtering mechanism rather than simply shortening the input text, we compare it with a random filtering strategy, denoted with the ``-rand'' suffix. The results are shown in Table \ref{tab: random_filter}. We observe that random filtering often performs worse than the original detector. This is expected because randomly removing subsequences does not distinguish hidden human-like spans from machine-indicative evidence, and may discard useful discriminative signals while merely reducing the text length.
This result supports our theoretical analysis that shortening the text alone can increase detection difficulty. Besides, it indicates that the improvement does not come from using shorter inputs, but from selectively reducing human-like spans, thereby validating the label-agnostic filtering strategy proposed in Section \ref{sec: theoretical_improvemenets}.

\subsection{Enhancement in a Training-free Manner}

The previous results evaluate the full proposed framework, including both the stacked architecture and the hard-EM-inspired optimization. We further consider a stricter setting, where an already trained detector is enhanced only through the stacked inference structure without any additional fine-tuning. Fig. \ref{fig: enhance_notrain} reports the results on ChatGPT-D, denoted as ChatGPT-D-No for the original detector and ChatGPT-STK-No for its training-free stacked version. Encouragingly, even without retraining, the stacked inference structure still brings clear improvements. This result shows that estimating and filtering confidently human-like subsequences at inference time can already help the detector focus on more machine-indicative evidence. Such a plug-and-play property makes the proposed framework flexible and scalable for practical applications. Besides, the overall gap between the filtering-only variant (Fig. \ref{fig: enhance_notrain}) and the full STK variant (Fig. \ref{fig: enhance_chatgpt}) measures the contribution of Section \ref{sec: stacked_framework}, since the only difference is whether the detector is further optimized with the latent retention masks.

\subsection{Complexity Evaluation}
\label{sec: running_time}

First, Table \ref{tab: running_time} compares the running time (training time and inference time) between the original detectors and the enhanced versions, fine-tuned for 5 epochs. The results are consistent with the complexity analysis discussed in Section \ref{sec: complexity}, indicating that the actual running time of the proposed enhancement framework does not exceed twice that of the original detector. We believe that achieving superior detection enhancement performance in an acceptable time frame is valuable.

Second, regarding memory usage, our stacked framework aims for efficient memory utilization. Specifically, the proposed framework uses the same detection model in both the E-step (filtering) and the M-step (final detection). Therefore, the peak memory usage is primarily determined by the detector model itself and is almost identical to the memory usage of the original benchmark detector, as no new model or parameters are introduced. Table \ref{tab: memory} also empirically validates this, showing that the additional memory introduced by the enhanced framework is negligible.

% Table generated by Excel2LaTeX from sheet 'memory'
\begin{table}[t]
\scriptsize
  \centering
  \caption{Memory consumption.}
  \begin{adjustbox}{width=0.5\textwidth}
  \setlength{\tabcolsep}{0.022\linewidth}{
    \begin{tabular}{c|ccc}
    \hline\noalign{\smallskip}
    Method & Essay & Reuters & SQuAD1 \\
    \noalign{\smallskip}\hline\noalign{\smallskip}
    ChatGPT-D/OpenAI-D/MPU & 9.36G & 9.36G & 9.15G \\
    \rowcolor{gray!20}\textbf{ChatGPT-STK/OpenAI-STK/MPU-STK} & 9.38G & 9.38G & 9.20G \\
    RADAR & 10.45G & 10.45G & 10.38G \\
    \rowcolor{gray!20}\textbf{RADAR-STK} & 10.47G & 10.47G & 10.42G \\
    \noalign{\smallskip}\hline
    \end{tabular}%
    }
    \end{adjustbox}
  \label{tab: memory}%
\end{table}%

\section{Conclusion}

This paper emphasizes the importance of the hidden human-like nature of machine-generated texts in MGT detection. Firstly, statistical analysis of existing datasets has empirically revealed the presence of hidden human-like spans, even in fully machine-generated texts. Then, we have theoretically demonstrated their negative impact on detection. Based on these theoretical findings, we have designed a stacked detection enhancement framework. Through theoretical analysis and extensive experimental evaluation, we have demonstrated the effectiveness of the proposed framework in enhancing existing detectors. Notably, our primary technical contribution lies in the conceptual framework, i.e., reducing the influence of hidden human-like spans in all texts, while the proposed hard-EM-inspired approach is one feasible implementation within this framework. This conceptual insight presents a promising direction for future work to further enhance MGT detection.

\bibliographystyle{IEEEtran}
\bibliography{ref}

% \clearpage
% \onecolumn
% \appendix
% \input{latex/appendix}

\end{document}